# b-Bit Minwise Hashing for Large-Scale Linear SVM

Ping Li
Cornell University
Ithaca, NY 14853
pingli@cornell.edu

Joshua L. Moore
Cornell University
Ithaca, NY 14853
jlmo@cs.cornell.edu

Arnd Christian König
Microsoft Research
Redmond, WA 98052
chrisko@microsoft.com

## ABSTRACT

Linear[1] Support Vector Machines (e.g., SVM$^{\text{perf}}$, Pegasos, LIBLINEAR) are powerful and extremely efficient classification tools when the datasets are very large and/or high-dimensional, which is common in (e.g.,) text classification. Minwise hashing is a popular technique in the context of search for computing resemblance similarity between ultra high-dimensional (e.g., $2^{64}$) data vectors such as document representations using higher-order shingles. b-Bit minwise hashing is a recent significant improvement over minwise hashing by storing each hashed value using only the lowest $b$ bits (instead of 64 bits).

In this paper, we propose to (seamlessly) integrate b-bit minwise hashing with linear SVM to substantially improve the training (and testing) efficiency using much smaller memory, with essentially no loss of accuracy. Theoretically, we prove that the resemblance matrix, the minwise hashing matrix, and the b-bit minwise hashing matrix are all positive definite matrices (kernels). However, since the resemblance kernel is non-linear, it appears not straightforward to use it for linear SVM. Interestingly, our proof for the positive definiteness of the b-bit minwise hashing kernel naturally suggests a simple strategy to integrate b-bit hashing with linear SVM, which only requires a very minimal modification of LIBLINEAR. Our technique is particularly useful when the data can not fit in memory, which is an increasingly critical issue in large-scale machine learning.

Our preliminary experimental results on a publicly available webspam dataset (350K samples and 16 million dimensions) verified the effectiveness of our algorithm. For example, the training time was reduced to merely **a few seconds**.

In addition, our technique can be easily extended to many other linear and nonlinear machine learning applications (on binary data) such as **logistic regression**. We will report

[1]First draft in Feb. 2011. Slightly modified in May 2011.

experimental results in subsequent manuscripts.

## 1. INTRODUCTION

The method of b-bit minwise hashing [30, 31, 33] is a very recent progress for efficiently (in both time and space) computing *resemblances* among extremely high-dimensional (e.g., $2^{64}$) binary vectors, which may be documents represented by $w$-shingles with $w = 5$ or 7 [2, 3]. In this paper, we show that b-bit minwise hashing can be seamlessly integrated with linear Support Vector Machine (SVM) [10, 15, 18, 37, 40]. In SIGKDD 2010, the nice work [40] addressed a critically important problem about training linear SVM when the data can not fit in memory. In this paper, our work also tackles the similar problem from a different dimension.

### 1.1 Minwise Hashing

The seminal work of *minwise hashing* [2, 3] has been successfully applied to a very wide range of real-world problems especially in the context of search, including duplicate Web page removal [2, 3], text reuse in the Web [1], detection of large-scale redundancy in enterprise file systems [12], syntactic similarity algorithms for enterprise information management [7], content matching for online advertising [36], Web graph compression [4], Web spam [17, 38], community extraction and classification in the Web graph [9], compressing social networks [8], advertising diversification [13], wireless sensor networks [19], graph sampling [35], and more.

Computing the size of set intersections is a fundamental problem in information retrieval, databases, and machine learning. For example, binary document vectors represented using $w$-shingles can be viewed either as vectors in very high-dimensions or as sets. Given two sets, $S_1$ and $S_2$, where

$$S_1, \ S_2 \subseteq \Omega = \{0, 1, 2, ..., D-1\},$$

a widely used measure of similarity is the *resemblance*

$$R = \frac{|S_1 \cap S_2|}{|S_1 \cup S_2|} = \frac{a}{f_1 + f_2 - a},$$

where $f_1 = |S_1|, \ f_2 = |S_2|, \ a = |S_1 \cap S_2|$.

Minwise hashing applies a random permutation $\pi : \Omega \to \Omega$ on $S_1$ and $S_2$. Based on an elementary probability result:

$$\mathbf{Pr}\left(\min(\pi(S_1)) = \min(\pi(S_2))\right) = \frac{|S_1 \cap S_2|}{|S_1 \cup S_2|} = R, \quad (1)$$

one can store the smallest elements under $\pi$, i.e., $\min(\pi(S_1))$ and $\min(\pi(S_2))$, and then repeat the permutation $k$ times to estimate $R$. After $k$ minwise independent permutations, $\pi_1, \pi_2, ..., \pi_k$, one can estimate $R$ without bias, as:

$$\hat{R}_M = \frac{1}{k}\sum_{j=1}^{k} 1\{\min(\pi_j(S_1)) = \min(\pi_j(S_2))\}, \quad (2)$$

$$\text{Var}\left(\hat{R}_M\right) = \frac{1}{k}R(1-R). \quad (3)$$

The common practice is to store each hashed value, e.g., $\min(\pi(S_1))$ and $\min(\pi(S_2))$, using 64 bits [11]. The storage cost (and consequently the computational cost) will be prohibitive in truly large-scale applications [34].

## 1.2 b-Bit Minwise Hashing

The recent development of *b-bit minwise hashing* [31] provides a solution to the (storage and computational) problem of *minwise hashing* by storing only the lowest b bits (instead of 64 bits) of each hashed value for a small b.

Again, consider two sets, $S_1, S_2 \subseteq \Omega = \{0, 1, 2, ..., D-1\}$. Define the minimum values under the random permutation $\pi : \Omega \to \Omega$ to be: $z_1 = \min(\pi(S_1))$ and $z_2 = \min(\pi(S_2))$.

Define $e_{1,i} = i$th lowest bit of $z_1$, and $e_{2,i} = i$th lowest bit of $z_2$. Theorem 1 provides an interesting probability result.

THEOREM 1. *[31] Assume D is large.*

$$P_b = \mathbf{Pr}\left(\prod_{i=1}^{b} 1\{e_{1,i} = e_{2,i}\} = 1\right) = C_{1,b} + (1 - C_{2,b})R$$

*where* $\quad r_1 = \frac{f_1}{D}, \quad r_2 = \frac{f_2}{D}, \quad f_1 = |S_1|, \quad f_2 = |S_2| \quad (4)$

$$C_{1,b} = A_{1,b}\frac{r_2}{r_1+r_2} + A_{2,b}\frac{r_1}{r_1+r_2}, \quad (5)$$

$$C_{2,b} = A_{1,b}\frac{r_1}{r_1+r_2} + A_{2,b}\frac{r_2}{r_1+r_2}, \quad (6)$$

$$A_{1,b} = \frac{r_1[1-r_1]^{2^b-1}}{1-[1-r_1]^{2^b}}, \quad A_{2,b} = \frac{r_2[1-r_2]^{2^b-1}}{1-[1-r_2]^{2^b}}. \square \quad (7)$$

Once the basic probability formula is known, one can repeat the permutations $k$ times to estimate $P_b$ in (4) from which one can estimate the resemblance $R$. That is

$$\hat{R}_b = \frac{\hat{P}_b - C_{1,b}}{1 - C_{2,b}}, \quad (8)$$

$$\hat{P}_b = \frac{1}{k}\sum_{j=1}^{k}\left\{\prod_{i=1}^{b} 1\{e_{1,i,\pi_j} = e_{2,i,\pi_j}\} = 1\right\}, \quad (9)$$

[31] carefully analyzed the variance of $\hat{R}_b$ and compared it with the variance of the original minwise hashing estimator $\hat{R}_M$. The result is encouraging. To estimate any $R \geq 0.5$, even in the least favorable situation, using $b = 1$ only requires to increase the number of permutations by a factor of 3, in order to achieve the same estimation variance as $\hat{R}_M$ (which uses $b = 64$ bits). Therefore, a 21.3-fold (64/3) improvement is attained if one is mainly interested in resemblance $R \geq 0.5$.

Interestingly, in this paper, we will show that we can use b-bit minwise hashing in linear SVM without directly using the estimator $\hat{R}_b$ (8). We will prove that the resemblance describes a family of positive definite kernels (hence naturally suitable for SVM), which however is *non-linear*. More interestingly, we will prove that the matrices generated by minwise hashing and b-bit minwise hashing are also positive definite. Furthermore, our proof directly suggests a simple implementation to use b-bit hashing with linear SVM, with only a very minimal modification of the original code.

## 1.3 Ultra High-Dimensional Large Datasets

In the context of search, a standard procedure to represent documents (e.g., Web pages) is to use $w$-shingles (i.e., $w$ contiguous words), where $w = 5$ or 7 in several studies [2, 3, 11]. This procedure can generate datasets of extremely high dimensions.

For example, suppose we only consider $10^5$ common English words. Using $w = 5$ may require the size of dictionary $\Omega$ to be $D = |\Omega| = 10^{25} = 2^{83}$; and $w = 7$ requires $D = 2^{117}$. In current practice, it looks $D = 2^{64}$ often suffices, as the number of available documents may not be large enough to exhaust the dictionary. However, as the Web continues to grow at a fast rate, it may be possible that we have to use $D > 2^{64}$ in the near future.

With $w \geq 5$, normally only the absence/presence (0/1) information is used, as a $w$-shingle is unlikely to occur more than once in a page. The total number of shingles is usually set to be $|\Omega| = 2^{64}$. Thus, the set intersection becomes the inner products in binary data vectors of $2^{64}$ dimensions. Interestingly, even when the data are not too high-dimensional, empirical studies [5, 14, 16] achieved good performance using SVM with binary-quantized (text or image) data.

The *webspam* dataset, which can be downloaded from the LibSVM site and was used in [40], is among the largest public classification datasets. It consists of 350,000 documents presented by 3-shingles in 16,609,143 dimensions. The average number of non-zeros per sample is about 3730. We will mainly use the *webspam* dataset to verify our proposed algorithm. We expect that the use of higher-order shingles (e.g., $w \geq 5$) in machine learning will become more common after we demonstrate the effectiveness of our algorithm which naturally integrates linear SVM with b-bit hashing.

## 2. LINEAR SVM

Linear SVMs have become very powerful and extremely popular. Representative software packages include SVM$^{\text{perf}}$ [18], Pegasos [37], and LIBLINEAR [10]. The *2008 PASCAL Large Scale Learning Challenge* compared various implementations of linear SVM and identified LIBLINEAR as the winner.

Given a dataset $\{(\mathbf{x}_i, y_i)\}_{i=1}^{n}$, $\mathbf{x}_i \in \mathbb{R}^D$, $y_i \in \{-1, 1\}$. SVM solves the following optimization problem (primal):

$$\min_{\mathbf{w}} \quad \frac{1}{2}\mathbf{w}^\mathbf{T}\mathbf{w} + C\sum_{i=1}^{n}\max\left\{1 - y_i\mathbf{w}^\mathbf{T}\mathbf{x_i}, \ 0\right\}, \quad (10)$$

where $C > 0$ is a penalty parameter. It is often more convenient to solve the dual problem:

$$\min_\alpha \quad f(\alpha) = \frac{1}{2}\alpha^\mathbf{T}\mathbf{Q}\alpha - \mathbf{e}^\mathbf{T}\alpha, \quad (11)$$
$$\text{subject to} \quad 0 \leq \alpha_i \leq C, \ i = 1, 2, ..., n.$$

where $Q_{ij} = y_i y_j \mathbf{x_i^T x_j}$ and $\mathbf{e} \in \mathbb{R}^n$ an all-one vector.

Our implementation will be based on the LIBLINEAR package [10], which implemented linear SVM using dual coordinate descent algorithm [15]; see Algorithm 1.

**Algorithm 1** The dual coordinate descent method for linear SVM [15]. We modified it for only $L_1$ linear SVM, which was considered in [40].

- Given $\alpha$ and $\mathbf{w} = \sum_i y_i \alpha_i \mathbf{x}_i$.
- While $\alpha$ is not optimal
  For $i = 1$ to $n$
  1. $\bar{\alpha}_i \leftarrow \alpha_i$
  2. $G = y_i \mathbf{w}^\mathbf{T} \mathbf{x}_i - 1$
  3.
  $$PG = \begin{cases} \min(G,0) & \text{if } \alpha_i = 0 \\ \max(G,0) & \text{if } \alpha_i = C \\ G & \text{if } 0 < \alpha_i < C \end{cases}$$
  4. If $|PG| \neq 0$,
  $$\alpha_i \leftarrow \min(\max(\alpha_i - G/Q_{ii}, 0), C)$$
  $$\mathbf{w} \leftarrow \mathbf{w} + (\alpha_i - \bar{\alpha}_i) y_i \mathbf{x}_i$$

## 2.1 The Memory Bottleneck

When the data can fit in memory, linear SVM is often extremely efficient after the data are loaded into the memory. It is however often the case that the data loading time dominates the computing time for solving the SVM problem [40].

A much more severe problem arises when the data can not fit in memory. This situation can be very common in practice. The publicly available *webspam* dataset needs about 24GB disk space, which exceeds the memory capacity of many desktop PCs. Note that *webspam* contains only 350,000 documents represented by 3-shingles. Practical applications, however, may involve hundreds of millions (or billions) of Web pages represented by (e.g.,) 5-shingles.

## 2.2 Block Linear SVM

[40] proposed solving the memory bottleneck by block linear SVM. Basically, they partitioned the data into $m$ blocks, calculated according to the available memory space. At each step, they loaded one block into the memory to update the coefficients $\alpha$ or $\mathbf{w}$ (which are assumed to reside in memory). When working with each block of data, they actually used the *Pegasos* [37] procedure internally.

While the block linear SVM algorithm provides a nice solution to the urgent practical problem, it does not appear to be very easy to implement. Moreover, the computational bottleneck is still at the memory because loading the data blocks for many iterations consumes a large number of disk IOs.

The authors of [40] conducted thorough experiments on three datasets. The largest dataset is *webspam*. Their experiments were conducted on a machine with only 1GB memory so that they could better investigate the impact of data splitting (such as block size) on the performance.

## 2.3 A Brief Introduction of Our Proposal

We propose a very different solution by using b-bit minwise hashing. We assume the data vectors are very high-dimensional and relatively very sparse. For example, if the dimension $D = 2^{64}$, then even a set of size $2^{54}$ (which corresponds to the equivalent of the amount of text in a small novel, when represented via shingles) will be relatively very sparse because $2^{-10} \approx 0.001$, even though in an absolute magnitude $2^{54}$ is a very large number.

We also consider that the data are binary, which as previously explained is a reasonable assumption in important practical scenarios. In fact, our experiments on *webspam* will show that binary quantizing that dataset essentially does not affect the accuracy.

With the above assumptions, we can apply b-bit minwise hashing on the dataset to obtain a very compact representation of the original data. Suppose we conduct $k$ permutations and store the lowest b bits for each hashed (i.e., the minimum) value, then the total storage is only $nbk$ bits.

For example, consider the *webspam* dataset ($n = 350000$), $b = 8$ and $k = 200$, then the total storage is only 70 MB. However, in order to use b-bit minwise hashing for linear SVM, we have to solve the following two problems:

- If we use b-bit minwise hashing to estimate the resemblance, which (we will soon prove) represents a positive definite *nonlinear* kernel, how can we effectively convert this nonlinear problem into a linear problem?

- We need to prove that the matrices generated by b-bit minwise hashing are indeed positive definite, which will provide the solid foundation for our proposed solution.

It turns out that our proof in the next section that *b-bit hashing matrices* are positive definite naturally provides the construction for converting the otherwise nonlinear SVM problem into linear SVM.

Clearly, one should notice that our method is not really a competitor of the approach in [40]. In fact, both approaches may work together to solve extremely large problems. For example, suppose $k = 200$ and $b = 8$, then one billion ($10^9$) documents may require 200GB memory, which may still exceed the capacity of most workstations. In this case, we can still apply the block linear SVM using the hashed data.

## 3. B-BIT MINWISE HASHING KERNELS

This section proves some theoretical properties of matrices generated by resemblance, minwise hashing, or b-bit minwise hashing. We will show that they are all positive definite matrices (kernels). Our proof not only provides a solid theoretical foundation for using b-bit hashing in SVM, but also illustrates the ideas behind the construction required for integrating linear SVM with b-bit hashing.

**Definition**: A symmetric $n \times n$ matrix $\mathbf{K}$ satisfying

$$\sum_{ij} c_i c_j K_{ij} \geq 0$$

for all real vectors $c$ is called *positive definite (PD)*. Note that here we do not differentiate PD from *nonnegative definite* following the convention in machine learning literature.

Consider $n$ sets $S_1, S_2, ..., S_n \in \Omega = \{0, 1, ..., D-1\}$. Apply one permutation $\pi$ to each set and define $z_i = \min\{\pi(S_i)\}$.

We will prove that the following three matrices are all PD.

- The *resemblance matrix* $\mathbf{R} \in \mathbb{R}^{n \times n}$, whose $(i, j)$-th entry is the resemblance between set $S_i$ and set $S_j$:
$$R_{ij} = \frac{|S_i \cap S_j|}{|S_i \cup S_j|} = \frac{|S_i \cap S_j|}{|S_i| + |S_j| - |S_i \cap S_j|} \quad (12)$$

- The *minwise hashing matrix* $\mathbf{M} \in \mathbb{R}^{n \times n}$:
$$M_{ij} = 1\{z_i = z_j\} \quad (13)$$

- The *b-bit minwise hashing matrix* $\mathbf{M}^{(b)} \in \mathbb{R}^{n \times n}$:
$$M_{ij}^{(b)} = \prod_{t=1}^{b} 1\{e_{i,t} = e_{j,t}\} \quad (14)$$

  where $e_{i,t}$ is the $t$-th lowest bit of $z_i$.

Our proof follows the basic principle. That is, a matrix $\mathbf{A}$ is PD if it can be written as an inner product $\mathbf{B}^{\mathbf{T}}\mathbf{B}$.

THEOREM 2. *The* minwise hashing matrix $\mathbf{M} \in \mathbb{R}^{n \times n}$ *defined by (13) is PD.*

**Proof:** *We can write*

$$M_{ij} = 1\{z_i = z_j\} = \sum_{t=0}^{D-1} 1\{z_i = t\} \times 1\{z_j = t\}$$

*Therefore, $M_{ij}$ is the inner product of two high-dimensional vectors of length $D$ and hence $\mathbf{M}$ can be written as an inner product $\mathbf{M} = \mathbf{B}^{\mathbf{T}}\mathbf{B}$, where $\mathbf{B} \in \mathbb{R}^{D \times n}$. This completes the proof.*□

THEOREM 3. *The* b-bit minwise hashing matrix $\mathbf{M}^{(b)} \in \mathbb{R}^{n \times n}$ *defined by (14) is PD.*

**Proof:** *We can write*

$$M_{ij}^{(b)} = \sum_{t=0}^{2^b-1} 1\{z_i = t\} \times 1\{z_j = t\}$$

*Therefore, $M_{ij}^{(b)}$ is the inner product of two $2^b$-dimensional vectors. This completes the proof.*□

THEOREM 4. *The resemblance matrix* $\mathbf{R} \in \mathbb{R}^{n \times n}$ *defined by (12) is PD.*

**Proof:** *The proof easily follows from the fact that $R_{ij} = \mathbf{Pr}\{M_{ij} = 1\} = E(M_{ij})$ and $M_{ij}$ is the $(i,j)$-th element of the PD matrix $\mathbf{M}$.*

One might be wondering if we need to worry about the fact that there are $k$ permutations instead of just one $\pi$. For example, there will be $k$ minwise hashing matrices: $\mathbf{M}_{(s)}$, $s = 1$ to $k$. Note that summation $\sum_{s=1}^{k} \mathbf{M}_{(s)}$ is still PD since

$$c^{\mathrm{T}}\left[\sum_{s=1}^{k} \mathbf{M}_{(s)}\right] c = \sum_{s=1}^{k} c^{\mathrm{T}} \mathbf{M}_{(s)} c \geq 0$$

for any vector $c$ by the fact that $\mathbf{M}_{(s)}$ is PD.

Similarly, the average $c^{\mathrm{T}}\left[\frac{1}{k} \sum_{s=1}^{k} \mathbf{M}_{(s)}\right] c \geq 0$.

Note that elements of a PD matrix satisfies the triangle inequality while the converse is not necessarily true. In fact, it is well-known that $R_{ij}$ satisfies the triangle inequality [2,6], although to the best of our knowledge, we have not seen a direct proof that the resemblance matrix $\mathbf{R}$ is PD.

On the other hand, the fact that $\mathbf{R}$ is PD does not seem to help us too much for efficient SVM training, because the resemblance is a nonlinear operation. However, the proof that the *b-bit minwise hashing matrix* $\mathbf{M}^{(b)}$ is PD provides us with a very simple strategy to construct a matrix $\mathbf{B}$ such that $\mathbf{M}^{(b)} = \mathbf{B}^{\mathbf{T}}\mathbf{B}$, where $\mathbf{B}$ has dimensions only $2^b \times 2^b$. As long as $b$ is not too large, this provides a highly affordable way to expand each minwise hashed value using $b$ bits.

## 4. INTEGRATING LINEAR SVM WITH B-BIT MINWISE HASHING

Using the construction in the proof of Theorems 2 and 3. our algorithm for integrating b-bit hashing with linear SVM becomes extremely simple (at least in retrospect).

Given a dataset $\{\mathbf{x}_i, y_i\}_{i=1}^{n}$, where $\mathbf{x}_i \in \mathbb{R}^D$ is a $D$-dimensional binary data vector (which is equivalent to a set). We apply $k$ independent random permutations on each $\mathbf{x}_i$ and store the lowest $b$ bits of each hashed value. This way, we obtain a new dataset which can be stored using merely $nbk$ bits. In the run-time, however, we need to expand each new data point into a $b \times k$-length vector.

For example, suppose $k = 3$ and the hashed values are originally $\{12013, 25964, 20191\}$, whose binary digits are $\{010111011101101, 110010101101100, 100110011011111\}$.

Consider $b = 2$. Then the binary digits are stored as $\{01, 00, 11\}$ (which corresponds to $\{1, 0, 3\}$ in decimals). In the run-time, we need to expand them into a vector of length $2^b k = 12$, to be

$$\{0, 0, 1, 0, \quad 0, 0, 0, 1, \quad 1, 0, 0, 0\}$$

which will be the new $\mathbf{x}_i$ fed to a linear SVM solver.

We have very slightly modified LIBLINEAR [10] to incorporate b-bit hashing. Since we really only need small $b$ values such as $b = 8$, both the training and testing become very efficient on the hashed dataset, as verified by experiments.

## 5. EXPERIMENTAL RESULTS ON WEBSPAM

Our experiment settings follow the work in SIGKDD 2010 [40] very closely. They conducted experiments on three datasets: (i) the *Yahoo-Korean* dataset is proprietary, (ii) the *epsilon* dataset is completely dense and low-dimensional, and (iii) the *webspam* is the largest among the three and reasonably high-dimensional ($n = 350000$, $D = 16609143$). Therefore, our experiments focus on the *webspam* dataset.

Following [40], we randomly selected 20% of samples for testing and used the rest 80% samples for training.

### 5.1 Binary v.s. Real-Value Data

Our current implementation is only for binary data, which is probably the most important case when documents are represented by $w$-shingles with $w \geq 5$. Even for *webspam* which only used $w = 3$, we notice that a binary-quantization on the dataset does not really affect the classification results.

Since there is a tuning parameter $C$, we conducted the experiments on a series of $C$ values ranging from 0.001 to 100. Figure 1 presents the results in terms of the number of support vectors (nSV), the testing accuracy (%), the training time (seconds), and the testing time (seconds), for both the original dataset and the binary-quantized dataset.

Note that, although we plot the results as functions of $C$ values, we do not intend to say that the results are directly comparable at a given $C$. There are two issues. Firstly, the two datasets will have different scales and hence their optimal $C$ values may be quite different. Secondly, in practice, we will conduct cross-validations to find the optimal $C$ for the best classifiers.

Since our purpose is to demonstrate the effectiveness of our proposed linear SVM scheme using b-bit hashing, we simply provide results for all $C$ values and we assume that the best performance is achievable if we conduct cross-validations.

Clearly, Figure 1 illustrates that binary-quantization on *webspam* does not degrade the performance in any aspect.

### 5.2 Evaluations of Testing Accuracy

Figures 2 (average) and 3 (std, standard deviation) provide the test accuracies. We experimented with $k = 30$ to $k = 500$, although prior practice [2,3,31] suggested that $k = 100$ to $k = 200$ should provide good results. We let $b = 1, 2, 4, 8,$ and 16.

Since our method is a randomized algorithm, we repeat every experiment 50 times. We report both the mean and std values. Figure 3 illustrates that the stand deviations are very small, especially with $b \geq 4$ ($< 0.1\%$). Figure 2 demonstrates that using $b \geq 8$ and $k \geq 100$ achieves about the same test accuracies as using the original data.

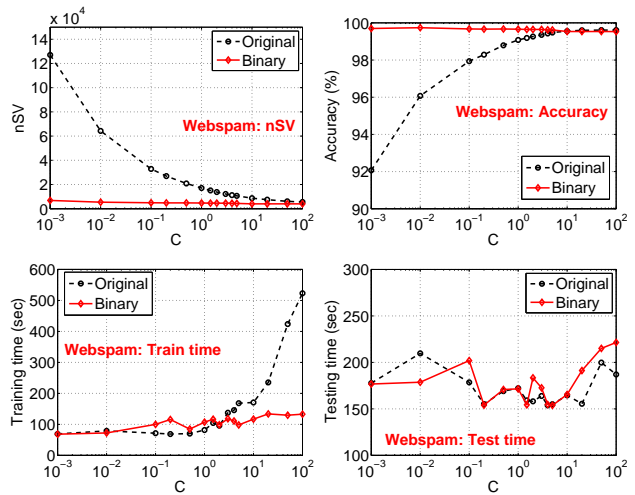

**Figure 1:** The purpose is to show that with binary quantization, the performance of linear SVM does not degrade. Note that we did not re-normalize the quantized data (to have unit norm) for this particular experiment. By private communications with Authors of LIBLINEAR, it looks they usually normalized the data, even for binary data. Therefore, our future reports will always normalize the data.

### 5.3 Evaluations of Training Time

Compared with the original training time (about 100 to 200 seconds in Figure 1), we can see from Figure 4 that only a very small fraction of the original cost is needed using our method.

Note that we did not include the data loading time in both the original method and our new method. Loading the original data took about 12 minutes while loading the hashed data took only about 10 seconds. Of course, there is a cost for processing (hashing) the data, which we find is very efficient, confirming prior studies [2]. In fact, data processing can be conducted during load collection, as the standard practice in search.

### 5.4 Evaluations of Testing Time

Compared with the original testing time (about 150 to 200 seconds in Figure 1), we can see from Figure 5 that only a very small fraction of the original cost is needed using our method (only a few seconds).

Note that the testing time includes both the data loading time and computing time, as designed by LIBLINEAR. The efficiency of testing may be very important in practice, for example, when the classifier is deployed in an user-facing application (such as search), while the cost of training or pre-processing (such as hashing) may be less critical and can often be conducted off-line.

## 6. RELATED WORK

In this paper, we focus on describing our method for significantly improving linear SVM in high-dimensional binary datasets, which are common in (commercial) text applications. As many integer data can be transformed into bi-nary data by (significantly) increasing the dimensions, our method is actually quite general. In fact, our method can be easily extended to many other linear and nonlinear learning algorithms such as **logistic regression**. We focus on linear SVM partly because other learning methods such as tree-based algorithms (which are also extremely popular in industry) are not particularly suitable for extremely high-dimensional (e.g., $2^{64}$) data. See one of the authors' recent work on ***abc-boost*** [22, 23], which had the detailed comparisons with (kernel) SVM and *deep learning* on a variety of not-too-high-dimensional datasets.

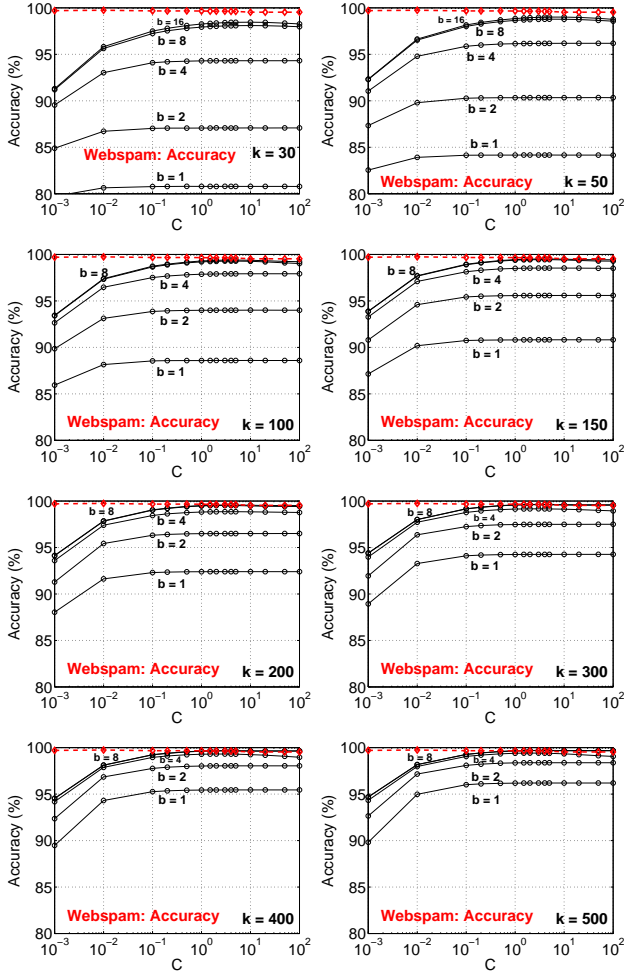

Figure 2: Test Accuracy. With $k \geq 100$ and $b \geq 8$, b-bit hashing achieves very similar accuracies as using the original (binary-quantized) data. The results are averaged over 50 repetitions.

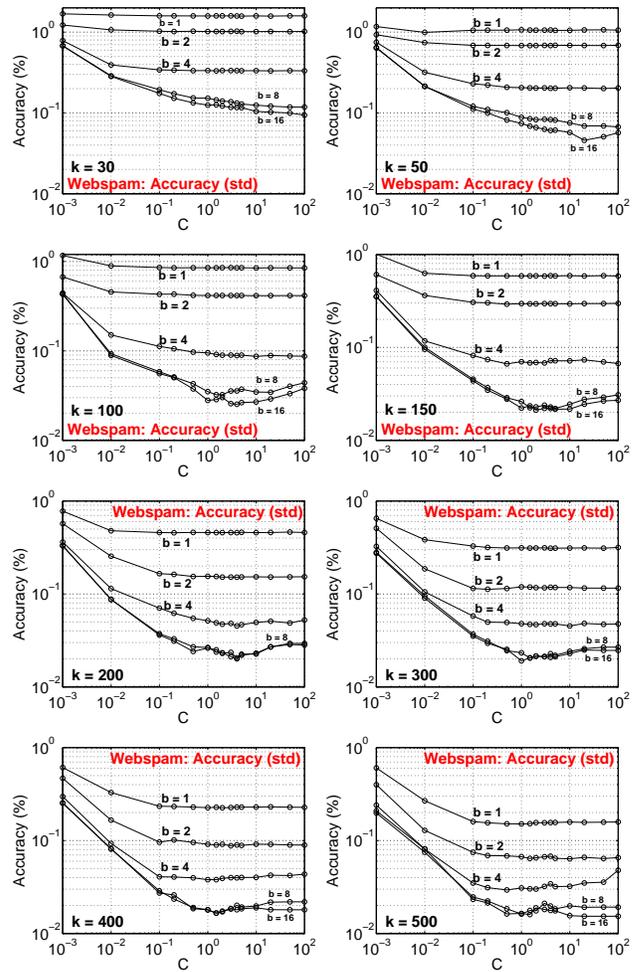

Figure 3: Test Accuracy (STD). The standard deviations are computed from 50 repetitions. When $b \geq 8$, the standard deviations become extremely small (e.g., $0.02\%$). This means our randomized algorithm produces very stable results.

In the past years, we have been working on a variety of hashing/sketching/sampling methods to deal with extremely large-scale high-dimensional data. Examples are *normal and normal-like random projections* [27, 28], *stable random projections* [20, 21, 26, 29], *Conditional Random Sampling (CRS)* [24, 25], as well as *b-bit minwise hashing* [30, 31, 33]. Many of those papers used (kernel) SVM as the motivating applications. Currently, we focus on *b-bit minwise hashing* because that method appears to be the state-of-the-art al-gorithm for binary data and now (as in this paper) we have discovered the simple and powerful technique to apply it to many large-scale learning problems.

Recently, a highly interesting hashing method was developed also for efficient SVM training [39], which reported the identical estimation variance as the special case in one of our earlier random projection papers [28] (by using "$s = 1$" in [28])[2]. Note that for randomized algorithms, it is essentially the variance which controls the storage size and algorithm complexity.

To compare *b*-bit minwse hashing with random projections (including [39], which reported the same variance as random projections [28]), a report [32] was written to compare their variances. [32] reported the theoretical comparisons, illustrating that *b*-bit minwise hashing improves random projec-tions often by 10- to 100-fold. In other words, to achieve the same accuracy, random projections would often require 10 to 100 times more storage than $b$-bit minwise hashing. This is a substantially large difference and should be noted by researchers and practitioners in large-scale machine learning.

---

[2] We appreciate John Langford, one of the authors of [39] for the highly helpful communications.

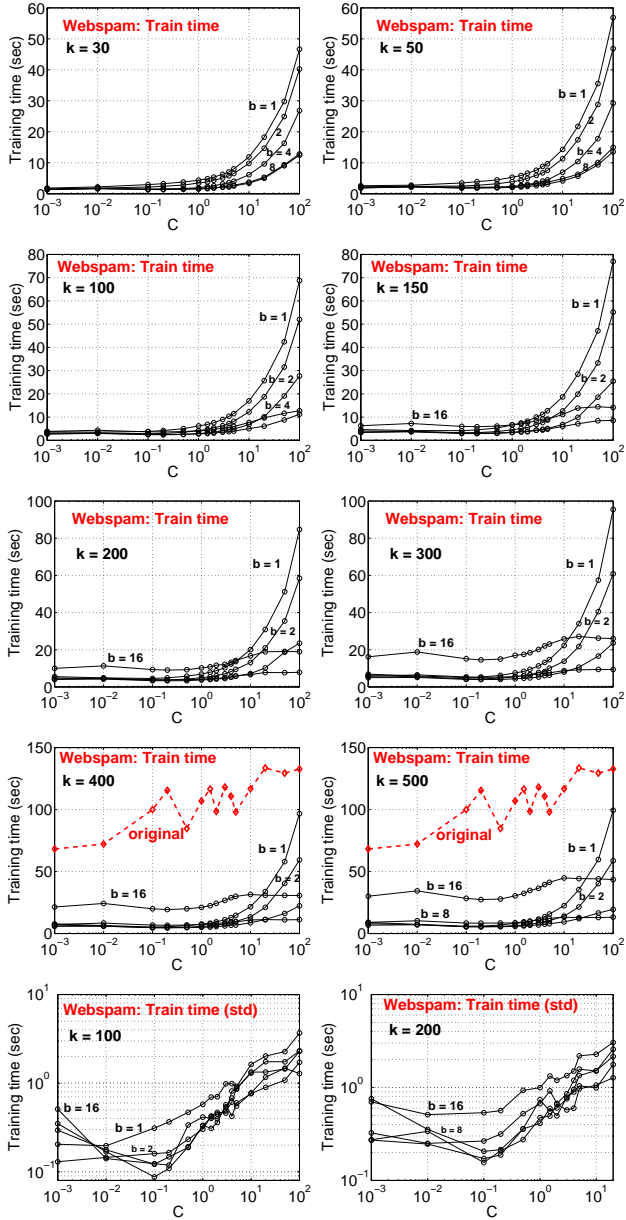

Figure 4: Training time. Compared with the training time of using the original data in Figure 1, we can see that our method with b-bit hashing only needs a very small fraction of the original cost. The bottom two panels plot the standard deviations.

## 7. CONCLUSION

We develop a simple and very efficient scheme to seamlessly integrate b-bit minwise hashing with linear SVM (in particular, LIBLINEAR). Our method requires only very small modification of the original code. Experiments demon-strate that our proposed method is very effective, using much smaller memory and less training time, to achieve essentially the same test accuracy. The testing stage also becomes much more efficient, which may be highly beneficial in important real-world applications such as search.

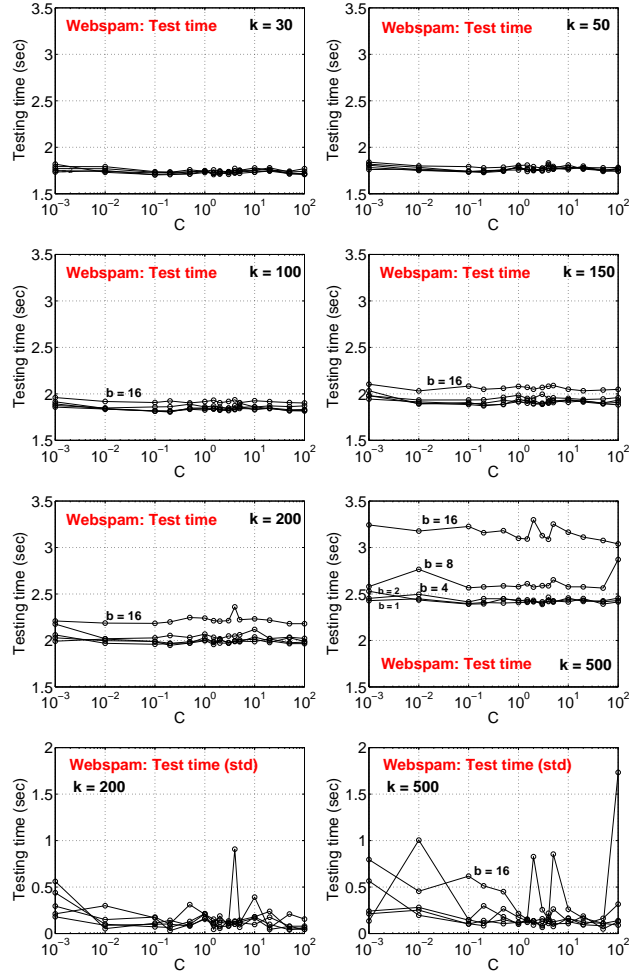

Figure 5: Testing time. Compared with the original testing time (about 150 to 200 seconds in Figure 1), only a very small fraction of the original cost is needed using our method. The bottom two panels plot the standard deviations.

Our proposed method provides an elegant and simple solution when the datasets do not fit in memory. In [40], they assumed the memory limit was only 1GB and hence they had to load the data in blocks (for multiple passes), incurring high IO costs. Note that, as we conducted our experiments on a workstation of 48 GB memory, we always loaded the entire dataset since the original (webspam) data size (about 24 GB) did not exceed the memory capacity. We expect that (commercial) applications may often involve high-dimensional datasets on the TB scale (or even much larger) and hence our method will have significant advantages for those applications.